\documentclass{article}

\PassOptionsToPackage{numbers, compress}{natbib}

\usepackage[final]{neurips_2025}

\usepackage[utf8]{inputenc} 
\usepackage[T1]{fontenc}    
\usepackage{hyperref}       
\usepackage{url}            
\usepackage{booktabs}       
\usepackage{amsfonts}       
\usepackage{nicefrac}       
\usepackage{microtype}      
\usepackage{xcolor}         

\usepackage{xcolor}
\usepackage{listings}
\lstdefinestyle{promptstyle}{
    backgroundcolor=\color{gray!10},
    basicstyle=\small\ttfamily,
    frame=single,
    xleftmargin=1em,
    xrightmargin=1em,
    framesep=4pt,
    rulecolor=\color{black!30},
    breaklines=true,
    columns=fullflexible,
    keepspaces=true
}

\definecolor{citecolor}{HTML}{2980b9}
\definecolor{linkcolor}{HTML}{c0392b}
\usepackage{hyperref}
\hypersetup{hidelinks,breaklinks=true,colorlinks,citecolor=citecolor,linkcolor=linkcolor}

\usepackage{overpic}
\usepackage{tabularx}
\usepackage{booktabs}
\usepackage{siunitx}
\usepackage{threeparttable}
\usepackage{multirow}
\newcommand{\tabincell}[2]{\begin{tabular}{@{}#1@{}}#2\end{tabular}}
\usepackage{amsmath}
\usepackage{amsfonts}
\usepackage{wrapfig}
\usepackage{enumitem}

\title{Align and Surpass Human Camouflaged Perception: Visual Refocus Reinforcement Fine-Tuning}

\author{
  \textbf{Ruolin Shen}$^{1}$,
  \textbf{Xiaozhong Ji}$^{2}$,
  \textbf{Kai Wu}$^{2}$,
  \textbf{Jiangning Zhang}$^{3}$,
  \textbf{Yijun He}$^{2}$, \\
  \textbf{HaiHua Yang}$^{2}$,
  \textbf{Xiaobin Hu}$^{1}$,
  \textbf{Xiaoyu Sun}$^{4}$ \\
  $^1$Technische Universität München\quad
  $^2$ByteDance  \quad 
  $^3$Zhejiang University \quad   
  $^4$Australian National University \quad  \\ 
  \vspace{-9mm}
}

\begin{document}

\maketitle

\centerline{\qquad \textbf{\color{magenta} Project Website}: \url{https://github.com/HUuxiaobin/VRRF}}

\begin{figure}[htbp]
\centering
\includegraphics[width=1\textwidth]{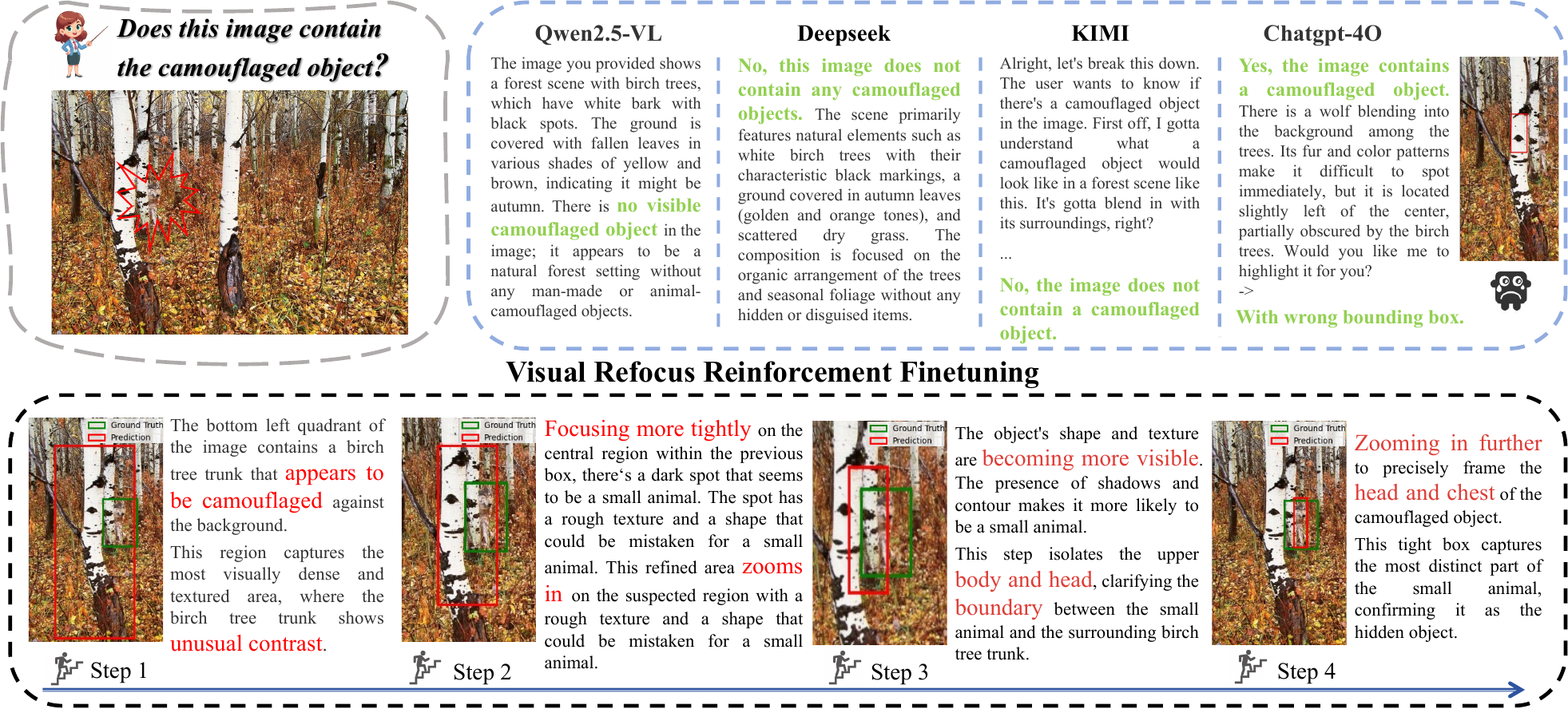}
\caption{ \footnotesize Intriguing discovery of SOTA multi-modal models on limitation: these models struggle to replicate human cognitive processes in leveraging foreground-background similarity relationships for visual analysis. Mimicking human visual camouflaged reasoning perception, our Visual Refocus Reinforcement Fine-Tuning visual system progressively and logically ‘refocus’ visual concealed content.
}
\label{fig:teaser}
\end{figure}

\begin{abstract}
Current multi-modal models exhibit a notable misalignment with the human visual system when identifying objects that are visually assimilated into the background. Our observations reveal that these multi-modal models cannot distinguish concealed objects,  demonstrating an inability to emulate human cognitive processes which effectively utilize foreground-background similarity principles for visual analysis. 
To analyze this hidden human-model visual thinking discrepancy, we build a visual system that mimicks human visual camouflaged perception to progressively and iteratively `refocus' visual concealed content. The refocus is a progressive guidance mechanism enabling models to logically localize objects in visual images through stepwise reasoning.  
The localization process of concealed objects requires hierarchical attention shifting with dynamic adjustment and refinement of prior cognitive knowledge. 
In this paper, we propose a visual refocus reinforcement framework via the policy optimization algorithm to encourage multi-modal models to think and refocus more before answering, and achieve excellent reasoning abilities to align and even surpass human camouflaged perception systems. 
Our extensive experiments on camouflaged perception successfully demonstrate the emergence of refocus visual phenomena, characterized by multiple reasoning tokens and dynamic adjustment of the detection box. Besides, experimental results on both camouflaged object classification and detection tasks exhibit significantly superior performance compared to Supervised Fine-Tuning (SFT) baselines. 
Furthermore,  our visual refocus system surpass the human camouflaged perception in user study where participants are required to identify the target object within a ten-second time limit — approximately equal to the model's average inference time.
We hope that this work will provide valuable insights for advancing future research in multimodal model development. The implementation code and datasets will be made publicly.  
\end{abstract}

\section{Introduction}
\label{sec:intro}
\vspace{-2mm}
Recent breakthroughs in applying reinforcement learning (RL) to large language models (LLMs) have yielded significant advancements. As demonstrated by OpenAI-O1 \cite{jaech2024openai}, reinforcement fine-tuning (RFT) strategies
exhibit substantial potential for enhancing LLMs' complex reasoning abilities. Furthermore, DeepSeek-R1-Zero \cite{guo2025deepseek} established that rule-based reward systems can effectively leverage RL to unlock superior reasoning and cognitive capabilities in language models, even without extensive supervised fine-tuning.

Recent research has increasingly sought to replicate the success of DeepSeek-R1 in multimodal large language models (MLLMs). Notably, Virgo \cite{du2025virgo} employed knowledge distillation from open-source reasoning models including DeepSeek-R1 \cite{guo2025deepseek} to enhance visual reasoning capabilities. However, the dominant research direction \cite{zhou2025r1, liu2025seg, zhan2025vision, deng2025boosting, peng2025lmm,liu2025visual,yang2025r1,zhang2025r1,deng2025openvlthinker} prioritizes direct implementation of DeepSeek-R1’s Group Relative Policy Optimization (GRPO) alongside its rule-based reward system to enable visual reasoning in MLLMs. 
This approach has primarily focused on improving performance in STEM tasks, particularly those involving mathematical reasoning and counting challenges. A fundamental driver of this success lies in the rule-based reward system's inherent applicability to tasks with deterministic ground-truth solutions, which enables both stable and interpretable reward signaling.
In the visual domain, many visually intensive tasks demand deeper cognitive processing and greater attention to visual information than reasoning with LLMs.
Through a comprehensive analysis, as illustrated in Fig. \ref{fig:teaser}, we identify a critical divergence between current multimodal models and human visual cognition in processing challenging camouflaged scenes. 
Specifically, these models fail to reliably detect objects visually assimilated into the background. Notably, even ChatGPT-4o exhibits hallucinations generating plausible explanations for potential concealment while ultimately failing to localize camouflaged objects.
Our findings indicate a critical limitation: their inability to detect visually concealed objects demonstrates a failure to emulate the human cognitive strategy of utilizing foreground-background similarity for visual interpretation. 

Motivated by these observations,  we are naturally led to explore whether rule-based reinforcement learning approaches can enhance the reasoning capability of Vision-Language Models (VLMs) to mimic the human perception system to iteratively refocus and refine the suspicious zone. 
We customize a visual `refocus' curriculum reinforcement learning to learn the visual refocus policy based on the rule-based reward design, which progressively learns a difficulty-hierarchically structured curriculum. Such hierarchical learning mechanism effectively mitigate the directional ambiguity caused by scalar rewards during model exploration.
Besides, we also tailor visual in-context refocus reinforcement learning paradigm to capture the context cognitive pattern. This paradigm guides the model in a stepwise manner, enhancing its logical reasoning ability, facilitating the emergence of refocusing capability, and improving both exploration flexibility and controllability.
After embedding such mechanisms, we observe the emergence of a "visual refocus" phenomenon, where the localization of concealed objects exhibits a hierarchical attention process that dynamically adjusts and refines prior cognitive representations. The ``visual refocus'' phenomenon mainly consists of three observable representation forms: including the `focus' (\textit{i.e.,} global to local zoom-in), `rethink' (\textit{i.e.,} local refinement and adjustment of suspicious zone) and `backtracing' (\textit{i.e.,} from local to global extension retracing).
Overall, the main contributions of this paper can be summarized as follows:
\noindent
\begin{itemize}[leftmargin=*]
\item We observe a notable misalignment between existing models and the human system on the perception of camouflaged objects, and propose Visual Refocus Reinforcement Fine-Tuning to align and surpass the human camouflaged perception ability. 

\item  We develop a visual `refocus' curriculum and in-context reinforcement learning paradigm,  respectively enabling hierarchical learning of task difficulty and capturing the context-aware cognitive patterns through a rule-based visual refocus reward. Such learning strategy triggers the ``visual refocus'' which iteratively and dynamically adjusts and refines prior cognitive knowledge.

\item 
We collect the camouflaged dataset and build the evaluation systems to analyze the camouflaged perception ability.
Through comprehensive experiments, we show that our approach to camouflaged object analysis achieves significantly better results than standard supervised fine-tuning (SFT) methods, especially on extremely challenging test set. Furthermore, empirical user studies demonstrate that our visual refocus system outperforms human behaviors in camouflaged object perception tasks.

\end{itemize}

\begin{figure}[t!]
\centering
\includegraphics[width=0.95\textwidth]{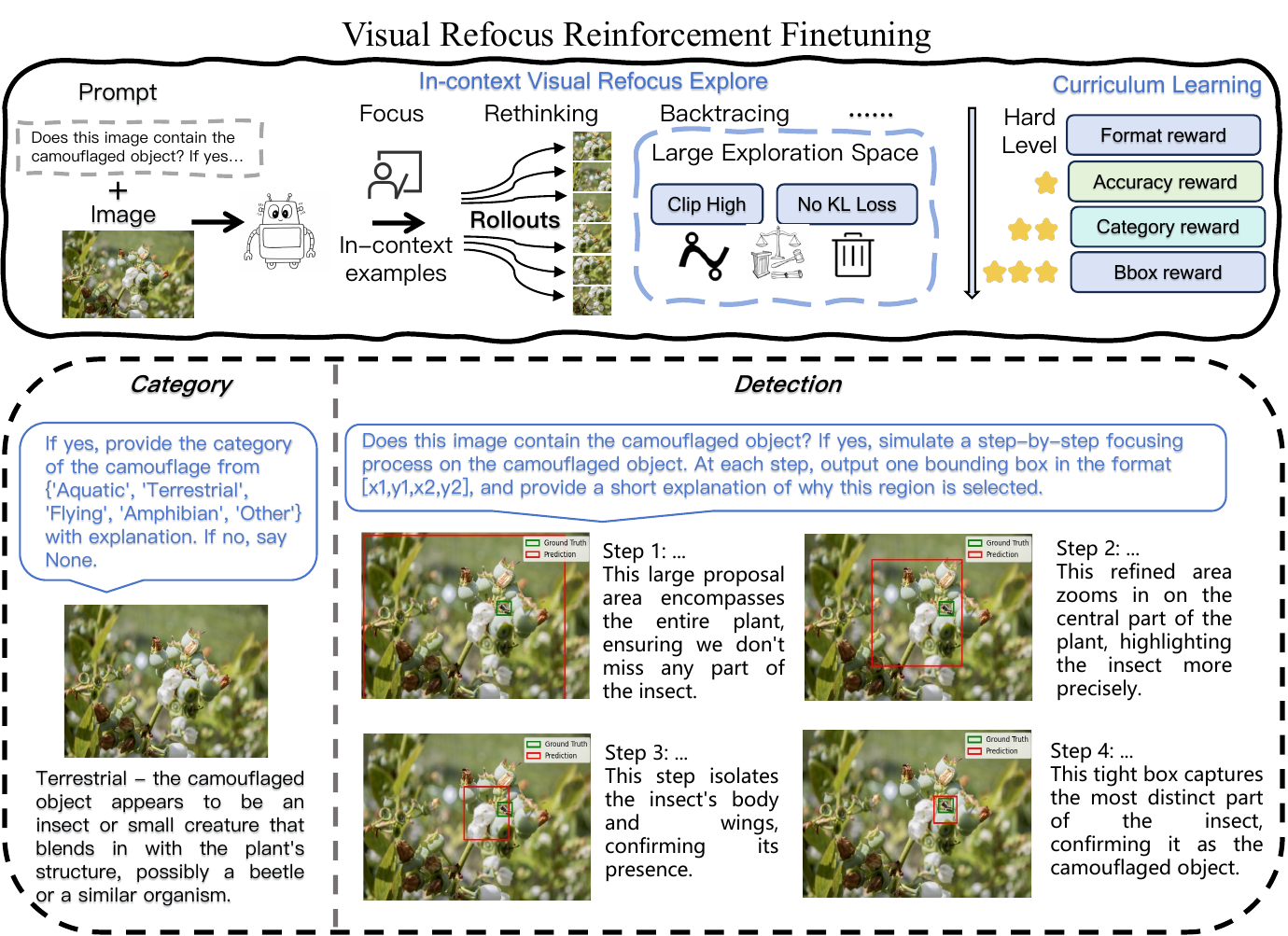}
\vspace{-3mm}
\caption{\footnotesize  \textbf{Overview of Visual Refocus Reinforcement Fine-Tuning}. 
}
\label{fig:llm_framework}
\vspace{-7mm}
\end{figure}

\vspace{-2mm}
\section{Related Work}
\vspace{-2mm}
\subsection{Vision-Language Models}
Vision-Language Models (VLMs) have witnessed remarkable development since the emergence of large language models (LLMs). Pioneering works such as Flamingo \cite{Alayrac2022} laid the foundation for VLMs, demonstrating their potential in few-shot learning tasks. Subsequently, LLaVA \cite{Liu2024b} employed GPT-4 \cite{Achiam2023} to generate training data, achieving promising results in visual dialogue and reasoning, which inspired a series of studies focusing on visual instruction data, like InstructBLIP \cite{Dai2023}. To address the limitation of constrained image input resolution in early VLMs, mechanisms like AnyRes \cite{Chen2024a,Chen2024c} were introduced, enabling flexible handling of images with different resolutions and aspect ratios and enhancing the models' perceptual and reasoning capabilities. Currently, popular open-source VLM series include LLaVA \cite{Li2024a,Liu2024b}, QwenVL \cite{Bai2025a,Wang2024a}, and InternVL \cite{Chen2024b,Chen2024c}. 
Building upon existing VLMs and inspired by the observed discrepancy between human and VLMs, we explore this interesting phenomenon and develop visual refocus system to better bridge and surpass the human camouflaged perception ability.


\vspace{-2mm}
\subsection{Reinforcement Learning in Vision-Language Models}
\vspace{-2mm}
The application of reinforcement learning (RL) in Vision-Language Models has become an active research area. DeepSeek R1 \cite{Guo2025} demonstrated that simple rule-based rewards can significantly enhance the reasoning capabilities of LLMs, inspiring researchers to explore the extension of similar RL methodologies to VLMs. Works like R1-OneVision \cite{Yang2025} proposed a cross-modal reasoning pipeline to improve VLM reasoning, while R1-V \cite{Chen2025a} introduced the GRPO method \cite{Shao2024} into VLM training for object-counting tasks. VisualThinker-R1-Zero \cite{Zhou2025} showed that applying RL to base VLMs can lead to substantial performance improvements and trigger the "visual aha moment".
Most of these previous studies on RL in VLMs target common visual understanding tasks or multimodal mathematics tasks. In contrast, 
our research focuses on the better visual reasoning alignment between humans and VLMs especially on the camouflaged object perception with unique challenges due to the nature of camouflaged objects blending into the background. 
Our visual refocus-based RL approach is designed to guide the model to perform a human-like chained thinking and visual focusing process, which is a novel way of developing a dynamical adjust
and refine prior cognitive representations. 



\vspace{-2mm}
\subsection{Camouflaged Object Perception}
\vspace{-2mm}
Camouflaged object perception, a bio-inspired research field, focuses on detecting concealed objects or animals that visually blend into their surroundings~\cite{fangtpami2021,hu2023high,tang2024chain}. Biological and psychological studies \cite{cuthill2019camouflage, stevens2009animal} demonstrate that camouflage serves as both a survival mechanism for prey species to evade predators and a perceptual challenge for human vision systems, which are particularly sensitive to edge-related color and illumination cues. Investigating camouflage phenomena offers valuable insights into the fundamental mechanisms of human visual perception.
Our method introduces multi-modal large models with RL and visual refocus technology, endowing the model with human-like high-level thinking and visual focusing abilities. This allows the model to not only analyze the visual features of the image but also conduct reasoning based on semantic information.

\label{sec:formatting}


\vspace{-2mm}
\section{Methods}
\vspace{-2mm}
In this section, we introduce our approach to stimulating the model's visual refocusing emergence abilities, enabling the model to progressively learn visual reasoning by adaptively switching and refining attention across global and local perspectives, guided by our proposed in-context GRPO framework with refinement.

\vspace{-2mm}
\subsection{Preliminary}
\vspace{-2mm}
\label{subsec:preliminary}
Group Relative Policy Optimization (GRPO)~\cite{shao2024deepseekmath} is a reinforcement learning algorithm designed to fine-tune large language models, particularly in scenarios with sparse or delayed rewards. GRPO simplifies traditional proximal policy optimization by eliminating the need for a separate value function estimator, thus reducing computational overhead.

In GRPO, for a given input (e.g., an image-question pair), the policy generates a group of $G$ outputs $\{o_1, o_2, \dots, o_G\}$. Each output $o_i$ receives a scalar reward $R_i$, possibly from a reward model or human feedback. The average reward for the group is computed as:
\(
\bar{R} = \frac{1}{G} \sum_{i=1}^G R_i.
\)
The relative advantage for each output is then:
\(
\hat{A}_i = (R_i - \bar{R}) / std(R).
\)
This approach focuses on the relative performance of each output within the group, promoting outputs that perform better than average.

The GRPO objective function is:
\begin{align}
L_{\text{GRPO}}^{\text{clip}}(\theta) = 
\mathbb{E}_{q \sim D,\; o_i \sim \pi_{\theta_{\text{old}}}(\cdot \mid q)} \Big[ \; &
\min\Big(
r_i(\theta) \hat{A}_i,\;
\mathrm{clip}(r_i(\theta), 1 - \epsilon, 1 + \epsilon) \hat{A}_i
\Big) \nonumber \\
& - \; \beta \cdot D_{\mathrm{KL}}\big[\pi_{\theta}(\cdot \mid q)\,\|\,\pi_{\text{ref}}(\cdot \mid q)\big]
\; \Big],
\end{align}
where $\beta$ controls the strength of the regularisation term, $q$ is the sampled question and $D_{\text{KL}}$ is the Kullback-Leibler divergence between the current policy and a reference policy.

While GRPO offers computational advantages, it has shown limited performance in tasks that require multi-step reasoning and fine-grained visual attention, such as camouflaged object detection. Our framework addresses these challenges by introducing a large exploration space with refocus priors, enabling more effective learning in complex visual environments.

\textbf{Problem Formulation}
Let \(x\!\in\!\mathcal{X}\) be an image containing a potentially camouflaged object, and let \(q\!\in\!\mathcal{Q}\) be an accompanying textual prompt (e.g., “Is there a hidden cat?”). Our model maintains an internal chain-of-thought (CoT) state \(h_t\) at step \(t\), which captures preceding “refocus” actions. We define:
\begin{equation}
s_t = \bigl(x,\,q,\,h_t\bigr), 
\quad 
a_t \sim \pi_\theta(a \mid s_t)
\label{eq:state_definition}
\end{equation}
where \(a_t\) is a discrete refocus instruction (e.g., “zoom into region \(R\)”). After \(T\) refocus steps, the model emits a final answer \(y\) via a read-out head:
\(
y \sim p_\theta\bigl(y \mid s_T\bigr).
\)
Our goal is to learn the policy parameters \(\theta\) that maximize the expected utility in identifying the concealed object.
\vspace{-2mm}
\subsection{Exploration-Aware Refocus Optimization}
\label{sec:exploration}
\vspace{-2mm}
To enable the model to reason through visually camouflaged scenes, our framework promotes broader and more flexible exploration in both the action and inference spaces. We incorporate two core strategies: in-context reinforcement learning for structured trajectory imitation, and a modified clip-high objective that encourages deviation from prior behavior without regularization penalties.

\paragraph{In-Context Reinforcement Learning with Trajectory Examples.}
Inspired by human learning processes, we design our reinforcement learning policy to facilitate exploration through illustrative examples. Regular GRPO encourages the model to explore autonomously; however, we observed that the performance ceiling is constrained by the model's inherent capabilities. To address this limitation, we integrate explicit in-context demonstrations of visual reasoning steps within each training example, structured specifically to guide and enhance the model's exploration process. We define the example exploration format as in Fig.~\ref{fig:prompt-example}, please check supplementary material for the complete prompt.

\begin{figure}[htbp]
    \centering
\begin{lstlisting}[style=promptstyle]
Does this image contain the camouflaged object?
<refocus instruction>
<format requirement>

# explore
<explore>
==== example i ====
Overview...
Focus (global to local zoom)...
Rethink (local refinement)...
Backtracing (local to global retracing)...
...
Summary.
==== example i+1 ====
...
==== example n ====
...
</explore>
# answers
<bbox>(x=112, y=98, w=64, h=52)</bbox>
<category>Camouflaged Category</category>
<answer>Yes</answer>
\end{lstlisting}
\vspace{-3.5mm}
    \caption{Prompt example used for in-context reinforcement learning. The \texttt{<explore>} block provides a multi-stage visual reasoning trajectory that mimics human perceptual shifts in attention.}
    \label{fig:prompt-example}
\end{figure}

Each exploratory example trajectory involves free-form attention adjustments (\textit{e.g.,} zooming, region recognition, refocus) to simulate the iterative process by which humans locate hidden elements. Rather than using structured or fixed exploration steps, we encourage flexible reasoning to enhance the model's capacity for adaptive thinking. This design allows the model to sequentially reason by adaptively attending to previously demonstrated exploratory behaviors, analogous to learning-by-demonstration approaches.
The in-context policy $\pi_\theta(a_t | h_{<t}, demo)$ conditions on prior steps $h_{<t}$ and the free-form demonstration. This formulation enables richer generalization without explicitly expanding the model size.

\vspace{-3mm}
\paragraph{Clip-High Objective Without KL Penalty.}
To encourage broader exploration and escape local optima under the instruction of in-context reinforcement learning, we revise the standard GRPO objective to a \textit{clip-high} variant with removing the KL divergence penalty and instead use a higher clipping ceiling.
We define the modified probability ratio as:
\(
r_i(\theta) = \frac{\pi_\theta(o_i \mid q)}{\pi_{\theta_{\text{old}}}(o_i \mid q)},
\)
and the new clip-high loss becomes:
\begin{align}
L_{\text{GRPO}}^{\text{clip-high}}(\theta) = 
\mathbb{E}_{q \sim D,\; o_i \sim \pi_{\theta_{\text{old}}}(\cdot \mid q)} \Big[ \;
\min\Big(
r_i(\theta) \hat{A}_i,\;
\mathrm{clip}(r_i(\theta), 1 - \epsilon, 1 + \delta) \hat{A}_i
\Big)
\; \Big],
\end{align}
where $\delta > \epsilon$ allows for a looser upper bound on policy shifts while still preventing collapse. By removing the KL term and expanding the upper clipping range, this formulation encourages the model to explore less likely (but potentially more optimal) paths and reduces the over-penalization of novel or rare reasoning trajectories.
We find that this modification leads to improved localization of highly camouflaged content and less reliance on prior policy or SFT biases.

\vspace{-2mm}
\subsection{Curriculum Reinforcement Learning for Progressive Reward Acquisition}
\vspace{-2mm}
In GRPO, the final reward is computed as the sum of multiple scalar rewards. This aggregation obscures detailed feedback on individual 
reward components, making it difficult to precisely discern which rewards are increasing or decreasing. Consequently, the model faces challenges in effectively learning the format or improving accuracy.
To further guide the model in mastering progressively harder aspects of camouflaged perception, we introduce a curriculum-style reinforcement learning schedule that incrementally augments the reward signal. Concretely, we define three stages—\emph{format \& accuracy}, \emph{category}, and \emph{localization IoU}—and successively incorporate them into the group relative policy optimization (GRPO) objective.

\paragraph{Stage 1: Format and Accuracy Rewards.}
In the initial phase, we focus the model on producing well-formed outputs and achieving basic recognition correctness. For each candidate output \(o_i\), we compute:
\begin{equation}
R_i^{(1)} = \lambda_{\mathrm{fmt}}\,R^{\mathrm{fmt}}(o_i) + \lambda_{\mathrm{acc}}\,R^{\mathrm{acc}}(o_i),
\end{equation}
where \(R^{\mathrm{fmt}}(o_i) \in [0,1]\) penalizes malformed XML tags or missing tokens in the \texttt{<bbox>}, \texttt{<category>}, and \texttt{<answer>} fields, \(R^{\mathrm{acc}}(o_i) \in \{0,1\}\) grants 1 point if the predicted \(y\) matches ground-truth presence, and \(\lambda_{\mathrm{fmt}}, \lambda_{\mathrm{acc}}\) are scaling coefficients.
We then compute the group-relative advantage \(\hat{A}_i^{(1)}\) using these \(R_i^{(1)}\) values exactly as in Section~\ref{subsec:preliminary}, and optimize the clip-high objective.

\paragraph{Stage 2: Adding Category Reward.}
Once the model reliably produces syntactically valid outputs and answers presence questions with the predefined format, we introduce semantic category correctness. Denote:
\begin{equation}
R^{\mathrm{cat}}(o_i) = 
\begin{cases}
1, & \text{if predicted category matches ground truth}, \\
0, & \text{otherwise}.
\end{cases}
\end{equation}

The cumulative reward becomes:
\begin{equation}
R_i^{(2)} = \lambda_{\mathrm{fmt}}\,R^{\mathrm{fmt}}(o_i) + \lambda_{\mathrm{acc}}\,R^{\mathrm{acc}}(o_i) + \lambda_{\mathrm{cat}}\,R^{\mathrm{cat}}(o_i),
\end{equation}
and we form the corresponding relative advantages \(\hat{A}_i^{(2)}\) from stage-2.

\paragraph{Stage 3: Incorporating IoU Refinement.}
Finally, to refine localization quality, we append an Intersection-over-Union (IoU) reward:
\begin{equation}
R^{\mathrm{iou}}(o_i) = \frac{\mathrm{area}(\mathrm{pred}_i \cap \mathrm{gt})}{\mathrm{area}(\mathrm{pred}_i \cup \mathrm{gt})},
\end{equation}
and define the full-stage reward:
\begin{equation}
R_i^{(3)} = \lambda_{\mathrm{fmt}}\,R^{\mathrm{fmt}}(o_i) + \lambda_{\mathrm{acc}}\,R^{\mathrm{acc}}(o_i) + \lambda_{\mathrm{cat}}\,R^{\mathrm{cat}}(o_i) + \lambda_{\mathrm{iou}}\,R^{\mathrm{iou}}(o_i).
\end{equation}
With \(\hat{A}_i^{(3)}\) computed in the usual way, our final curriculum objective is:
\begin{align}
L^{(3)}(\theta) = \mathbb{E}_{q, o_i} \left[ \min\left( r_i(\theta)\,\hat{A}_i^{(3)},\, \mathrm{clip}(r_i(\theta), 1 - \epsilon, 1 + \delta)\,\hat{A}_i^{(3)} \right) \right].
\end{align}

We transition from one stage to the next when the reward ceases to increase, which typically occurs after approximately 2--6 epochs. In practice, we set $\lambda_{\mathrm{fmt}} = \lambda_{\mathrm{acc}} = \lambda_{\mathrm{cat}} = \lambda_{\mathrm{iou}} = 1$ without tuning the trade-off parameters. 
This progressive pipeline enables the model to first master the output structure and basic presence awareness, then semantic classification, and ultimately fine-grained localization. By structuring reward signals into such a curriculum, our framework guides the model through increasingly complex visual reasoning tasks---mirroring the progressive ``refocusing'' characteristic of human perception---while preserving the computational efficiency of GRPO's relative advantage formulation.

\section{Experiments}

\subsection{Experimental Setup}
\noindent \textbf{Implementation Details.}
We use Qwen-2.5-VL-7B \cite{bai2025qwen2} as the base model. The default GRPO algorithm settings are adopted, with the number of generations N=4 and temperature set to 1. We train the model using the AdamW optimizer, starting with a learning rate of 1e-6, which is linearly decayed over the course of training. The model is trained for 2 epochs with a total batch size of 8. Training is completed within approximately one day using 8 NVIDIA H20 GPUs.

\begin{wrapfigure}{l}{0.48\textwidth}
\includegraphics[width=1\linewidth]{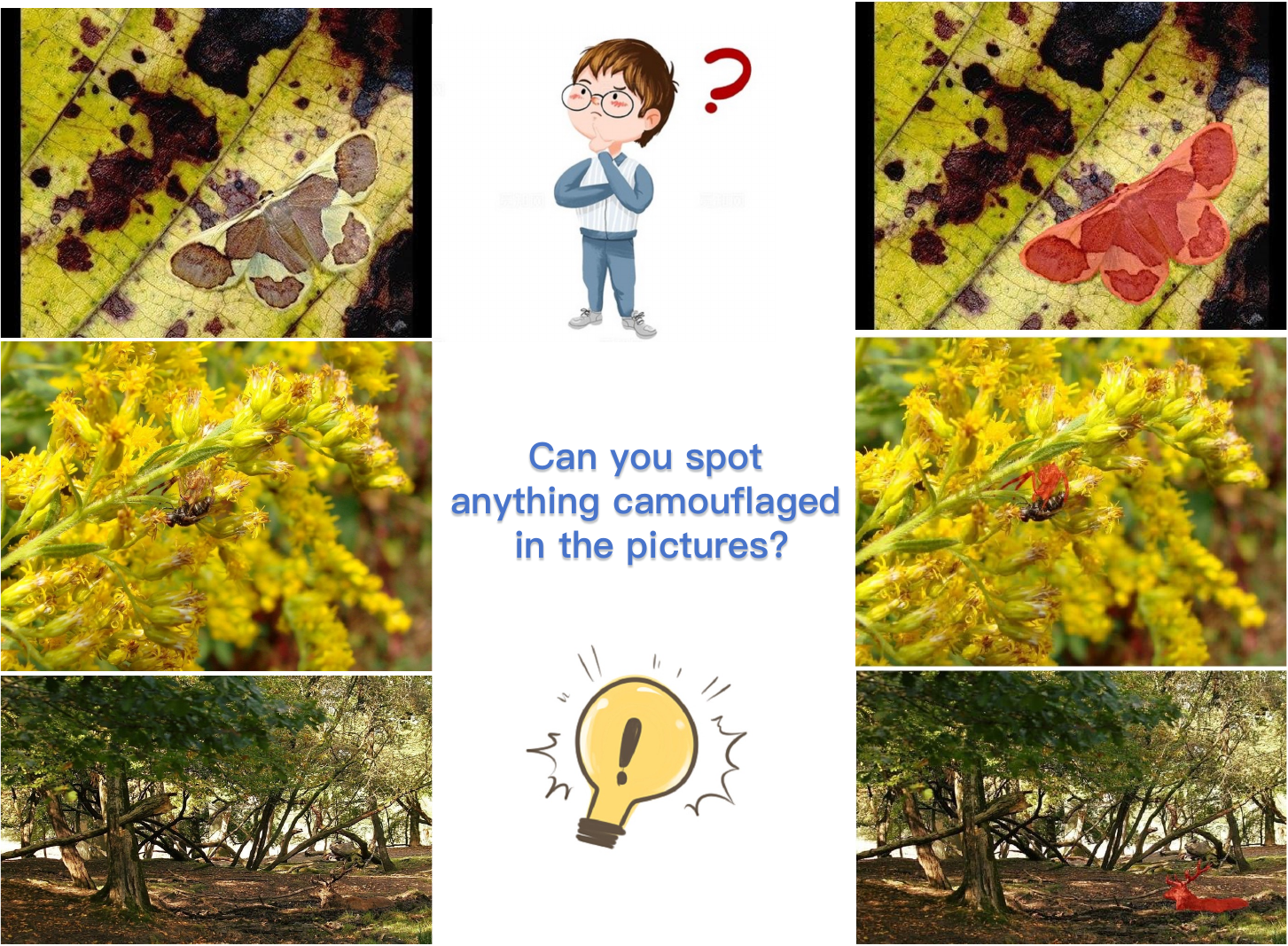}
\vspace{-3mm}
\caption{\footnotesize{Examples from our hard-concealed object set. Can you find them? Best viewed in color and zoomed-in.}}
\label{fig:cod_set}
\vspace{-3mm}
\end{wrapfigure} 
\noindent \textbf{Datasets and Evaluation Metrics.}
We evaluate our proposed VRRF on four public benchmark datasets for Camouflaged Object Detection (COD): COD10K \cite{fan2020camouflaged}, NC4K \cite{lv2021simultaneously}, CAMO \cite{le2019anabranch} and CHAMELEON \cite{2018Animal}. These datasets cover a wide range of challenging camouflage scenarios.
To assess model performance under varying levels of difficulty, we construct two subsets: an easy and a hard concealed object test set as shown in Fig. \ref{fig:cod_set}. The hard test set is extremely challenging and 
is manually selected from the full datasets. The easy set 
is sampled images from the remaining data.
In total, our experimental data includes 14,017 training samples (comprising 9,083 camouflaged and 4,934 noncamouflaged).
For the category classification task, we adopt the five super-categories defined in the COD10K dataset: Aquatic, Terrestrial, Flying, Amphibian, and Other. The model is explicitly instructed to choose from these predefined classes to ensure consistent labeling. 
For the detection task, since all datasets provide segmentation masks, we derive bounding boxes from the available masks. If multiple camouflaged objects are present in an image, the model is only required to predict one bounding box per image, capturing any of the camouflaged objects present. To evaluate the classification task, we use the following metrics: Category Accuracy, weighted precision, weighted recall, and weighted F1 score. These provide complementary perspectives on the classification performance across all categories. For the detection task, we report: Mean Intersection over Union (mIoU), IoU$@$0.3, IoU$@$0.5, and IoU$@$0.7 thresholds as well as mean center distance. These metrics measure the overlap quality between predicted and ground truth bounding boxes, capturing both coarse and fine-grained localization performance.

\vspace{-2mm}
\subsection{Camouflaged  Object Perception and Category Classification}
\vspace{-2mm}
As shown in Tab. \ref{tab:tabhard}, on the concealed challenging hard dataset,  even humans struggle to recognize these difficult camouflaged cases within seconds. 
Our method outperforms Supervised Fine-Tuning (SFT) on the category classification and concealed object perception primarily by modeling the reasoning process rather than merely supervising the final output. Our Visual Refocus Policy training paradigm guides the model to progressively localize objects through intermediate bounding box predictions (\textit{e.g.,} focusing on discriminative parts like animal heads/chests/wings). This process-oriented approach, combining step-by-step visual grounding with classification, significantly improves final classification and perception accuracy. 

\vspace{-2mm}
\subsection{Camouflaged Object Detection}
\vspace{-2mm}
On the hard concealed test set of Tab. \ref{tab:tabhard}, the performance gain of VRRF is significantly greater than the increase on the easy concealed test set in Tab. \ref{tab:tab1}. It indicates that tasks of higher difficulty require stronger reasoning ability and gain substantial improvements from a well-equipped reasoning model. Notably, our system even surpasses human performance in the hard cases, as confirmed by user studies where human participants struggled to accurately localize and categorize the camouflaged objects within a limited time frame. The superior performance is primarily attributed to our novel mechanism - an integrated framework combining reasoning process, joint training, and refocusing capability, which enables progressive localization of camouflaged objects in complex environments through iterative reasoning.

\begin{table*}[t!]
  \centering
    \small 
  \caption{\small 
    Quantitative evaluation results on Easy-Concealed Object test set, decomposed into camouflaged object classification and detection tasks. The best results are highlighted in \textbf{bold} and the second-best is marked in \underline{underline}.
  }\label{tab:tab1}
  \renewcommand{\arraystretch}{1} 
  \resizebox{1\textwidth}{!}{
  \begin{tabular}{l||c|cccc}
  \toprule
Easy Concealed& \multicolumn{1}{c|}{\tabincell{c}{Concealed Object Existence}}  & \multicolumn{4}{c}{\tabincell{c}{Concealed Category Classification}}   \\
    Object Set~~~ & Binary Acc $\uparrow$ & Category Acc  $\uparrow$ &  Precision $\uparrow$ & Recall $\uparrow$  & F1 $\uparrow$       \\
    \midrule
    GPT4.1 ~\cite{Achiam2023} & \underline{0.99} &	0.82 &0.87& 0.82 & 0.81  \\

    Qwen2.5-vl-7B ~\cite{Bai2025a} & 0.84 &	0.56 &0.53& 0.56 & 0.52  \\
    Qwen2.5-vl-72B ~\cite{Bai2025a} & 0.95 &	0.65 &0.80& 0.65 & 0.63  \\
     \midrule
    R1-V ~\cite{Chen2025a}  &0.97	& 0.63 & 0.50  & 0.63 & 0.52 \\
    SFT  &0.97	& \underline{0.84} & \underline{0.89}  &\underline{0.84} & \underline{0.86} \\
    VRRF  &\textbf{0.99}	& \textbf{0.89} & \textbf{0.90}  &\textbf{0.89} & \textbf{0.90} \\
  \toprule
 Easy Concealed & \multicolumn{5}{c}{\tabincell{c}{Concealed Detection}}   \\
      Object Set & mIOU $\uparrow$ & IoU $\geq$ 0.3(\%)  $\uparrow$ &  IoU $\geq$  0.5(\%) $\uparrow$ & IoU $\geq$ 0.7(\%) $\uparrow$  & Mean center distance(px) $\downarrow$       \\
      \midrule
    GPT4.1 ~\cite{Achiam2023} & 0.406 &	66.67 &37.25& 8.82 & 141.17  \\
    Qwen2.5-vl-7B ~\cite{Bai2025a} &0.244 &32.35 &20.59 & 13.73 & 93.90  \\
    Qwen2.5-vl-72B ~\cite{Bai2025a} & 0.467 &	69.61&48.04& 25.49 & \underline{57.72} \\
     \midrule
    R1-V ~\cite{Chen2025a}  & 0.496	& 69.61 & 51.96  &\underline{35.29} & 162.75 \\
    SFT  &\underline{0.528}	& \underline{81.37} & \underline{58.82}  &29.41 & 110.96 \\
    VRRF  &\textbf{0.726}	& \textbf{92.16} & \textbf{82.35}  &\textbf{67.65} & \textbf{35.98} \\
  \toprule
  \end{tabular}
 }
  \vspace{-5mm}
\end{table*}   

\begin{table*}[t!]
  \centering
    \small 
  \caption{\small 
    Quantitative evaluation results on Hard-Concealed Object test set, decomposed into camouflaged object classification and detection tasks. The best results are highlighted in \textbf{bold} and the second-best is marked in \underline{underline}.
  }\label{tab:tabhard}
  \renewcommand{\arraystretch}{1} 
  \resizebox{1\textwidth}{!}{
  \begin{tabular}{l||c|cccc}
  \toprule
Hard Concealed& \multicolumn{1}{c|}{\tabincell{c}{Concealed Object Existence}}  & \multicolumn{4}{c}{\tabincell{c}{Concealed Category Classification}}   \\
    Object Set~~~ & Binary Acc $\uparrow$ & Category Acc  $\uparrow$ &  Precision $\uparrow$ & Recall $\uparrow$  & F1 $\uparrow$       \\
    \midrule
    Human perception &0.654 &	0.46 &0.78 & 0.46 & 0.58 \\
    \midrule
    GPT4.1 ~\cite{Achiam2023} & \underline{0.913} & \underline{0.78} & 0.85& \underline{0.78} & \underline{0.79}  \\
    Qwen2.5-vl-7B ~\cite{Bai2025a} &0.423 &	0.28 &0.73 & 0.28 & 0.38  \\
    Qwen2.5-vl-72B ~\cite{Bai2025a} & 0.500 &	0.44 &\textbf{0.89}& 0.44 & 0.58 \\
    \midrule
    R1-V ~\cite{Chen2025a}  & 0.789	& 0.59 & 0.74  & 0.59 & 0.62 \\
    SFT  &0.830	& 0.70 & 0.86  &0.70 & 0.75 \\
    VRRF  &\textbf{0.923}	& \textbf{0.80} & \underline{0.88}  &\textbf{0.80} & \textbf{0.82} \\
  \toprule
 Hard Concealed & \multicolumn{5}{c}{\tabincell{c}{Concealed Detection}}   \\
      Object Set & mIOU $\uparrow$ & IoU $\geq$ 0.3(\%)  $\uparrow$ &  IoU $\geq$  0.5(\%) $\uparrow$ & IoU $\geq$ 0.7(\%) $\uparrow$  & Mean center distance(px) $\downarrow$       \\
      \midrule
      Human perception &\underline{0.382} &49.04	 &\underline{42.31} & \underline{32.69} & 69.82  \\
      \midrule
      GPT4.1 ~\cite{Achiam2023} & 0.239 & 40.38 &14.42& 3.85 & 143.88  \\
      Qwen2.5-vl-7B ~\cite{Bai2025a} &0.120 &15.38	 &10.58 & 7.69 & 83.30  \\
      Qwen2.5-vl-72B ~\cite{Bai2025a} & 0.249 &33.65 &25.00& 15.38 & \textbf{55.16} \\
    \midrule
    R1-V ~\cite{Chen2025a} & 0.286	& 40.38 & 26.92  &15.38 & 135.17 \\
    SFT  &0.365	& \underline{53.85} & 37.50  &20.19 & 89.98 \\
    VRRF  &\textbf{0.473}	& \textbf{60.58} & \textbf{54.81}  &\textbf{38.46} & \underline{61.11} \\
  \toprule
  \end{tabular}
 }
  \vspace{-5mm}
\end{table*}

\begin{table*}[t!]
  \centering
    \tiny
  \caption{\small 
  Qualitative results of the ablation study on the Hard-Concealed Object test set. The best results are highlighted in \textbf{bold}.
  }\label{tab:ablation_study}
  \renewcommand{\arraystretch}{1} 
  \resizebox{1\textwidth}{!}{
  \begin{tabular}{l||c|cc|ccc}
  \toprule
Hard Concealed& Concealed Object Existence  & \multicolumn{2}{c|}{\tabincell{c}{Classification}} & \multicolumn{3}{c}{\tabincell{c}{Detection}}   \\
    Object Set~~~ & Binary Acc  & Category Acc   &  F1 & mIOU   & IoU $\geq$ 0.5(\%)  & Mean center distance     \\

    \midrule
    Qwen2.5-vl ~\cite{Bai2025a} &0.423 &	0.28 &0.38 & 0.120 & 10.58 &83.30 \\
    + format \& acc reward &0.789 &	- &- & - & - &-  \\
    + category reward &0.837	& 0.73 & 0.79  &- & - & -\\
    + bbox reward  &0.856	& 0.74 & 0.79  &0.420 & 42.31 & 82.72 \\
    + in-context learning & \textbf{0.923} & \textbf{0.80} & \textbf{0.82}  & \textbf{0.473} & \textbf{54.81} & \textbf{61.11} \\
     \midrule
    All-at-once (No Curriculum)  & 0.528 & 0.45 & 0.59  & 0.310 & 36.54 & 61.17 \\
  \toprule
  \end{tabular}
 }
  \vspace{-5mm}
\end{table*}

\begin{figure}[htbp]
\centering
\includegraphics[width=0.95\textwidth]{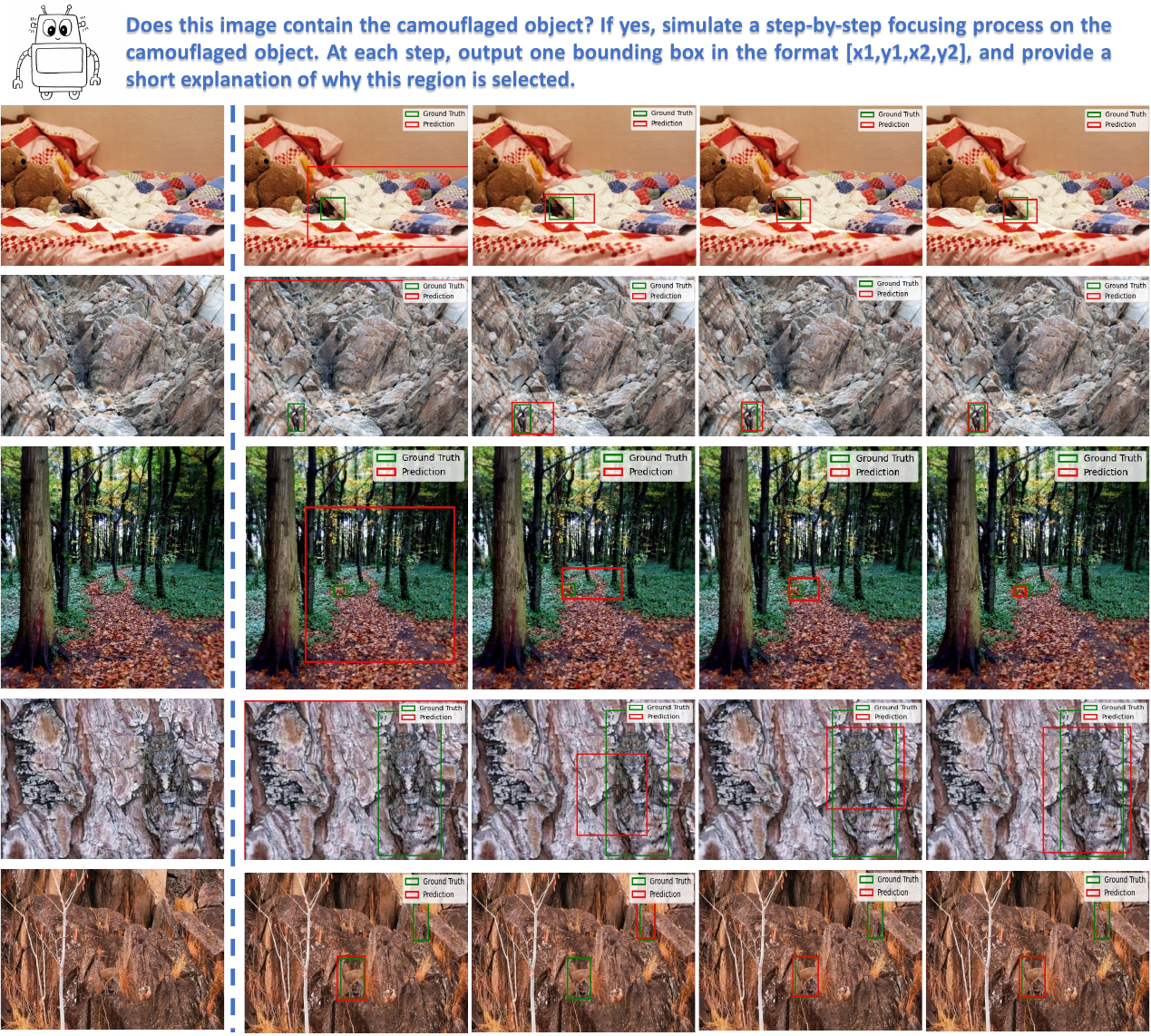}
\vspace{-3mm}
\caption{\footnotesize \textbf{Illustration of ``Visual Refocus'' representation pattern}. The first three rows show `focus' in the form of global to local zoom-in, 4-th row denotes `backtracing' from local to global extension retracing after perceiving the discriminate head and wing part, and 5-th row means `rethink' to refine and adjust the box to detect other same objects. 
}
\label{fig:visual_refocus}
\vspace{-7mm}
\end{figure}
\vspace{-2mm}
\subsection{Visual Refocus}
\vspace{-2mm}
Our visual refocus mainly demonstrate three refocus paradigms including the `Focus' (\textit{i.e.,} global to local zoom-in), `Rethink' (\textit{i.e.,} local refinement) and `Backtracing' (\textit{i.e.,} local to global focus). Specifically, during the stepwise localization process, the initial bounding box typically adopts a coarse-grained scope to ensure comprehensive coverage of potential clues. 
As a `Focus' mechanism,  subsequent steps progressively focus with the reasonable tokens (\textit{i.e.,}, zoom in) to capture finer details (\textit{e.g.,} head/chest) until converging to the final precise box.
Notably, our framework supports `Backtracing'  refinement paradigms. For instance, as shown in Fig. \ref{fig:visual_refocus} - 4-th row, the model may directly attend to isolate discriminative local features of the cicada at intermediate steps, such as its head and wings, which stand out due to their distinct shape and color, and then zoom out to capture the whole object.
Additionally, in multi-instance scenarios where only one target out is required, the model develops the `Rethink' ability to refine the bounding-box: it may detect other similar objects at intermediate steps but ultimately re-focuses on the primary target to deliver a single refined output box.

\subsection{Ablation Studies}
As shown in Tab. \ref{tab:ablation_study}, we conduct extensive ablation studies progressively adding the refocus policy including the format, accuracy, category and bbox reward mechanism, and the in-context learning reinforcement learning. We observe that the addition of each reward will gradually boost both the classification accuracy and the detection performance. Such joint training framework enhances the model’s reasoning capability by enabling iterative refinement. In-context learning further guides the model on how to think by teaching it to dynamically refocus attention to detect camouflaged objects, verify intermediate results, and progressively refine predictions (\textit{e.g.,} through refocus adjustments based on prior outputs). In contrast, adding all components simultaneously (\textit{i.e.,} no curriculum learning) results in inferior performance, likely due to the complexity of jointly optimizing the camouflaged object perception tasks, and therefore highlighting the benefit of our progressive learning.

\section{Conclusion}
In this paper, we observe that current multimodal models fail to detect concealed objects, lacking the human cognitive ability to analyze foreground-background similarity. We address the critical gap between multimodal models and human cognition in detecting camouflaged objects by introducing a visual refocus reinforcement framework. By emulating human-like hierarchical reasoning to progressively shift attention, our method enables models to dynamically refine and refocus their predictions through stepwise analysis. Extensive experiments demonstrate that this approach not only outperforms SFT baselines in classification and detection tasks but also exhibits emergent human-aligned refocusing behaviors, characterized by multi-token reasoning. Our work provides a pathway toward more cognitively inspired multimodal systems.

\textbf{Limitations and Broader Impacts}
Our proposed visual refocus reinforcement learning framework effectively enhances reasoning and cognitive capabilities by prompting MLLMs to refocus attention before answering. While our method can positively impact applications like autonomous driving, medical image analysis, and security surveillance through improved visual recognition, it also risks misuse in privacy-invasive surveillance systems, highlighting ethical considerations for responsible deployment.

{
    \small
    \bibliographystyle{unsrt}
    \bibliography{neurips_2025}

\begin{thebibliography}{10}

\bibitem{jaech2024openai}
Aaron Jaech, Adam Kalai, Adam Lerer, Adam Richardson, Ahmed El-Kishky, Aiden Low, Alec Helyar, Aleksander Madry, Alex Beutel, Alex Carney, et~al.
\newblock Openai o1 system card.
\newblock {\em arXiv preprint arXiv:2412.16720}, 2024.

\bibitem{guo2025deepseek}
Daya Guo, Dejian Yang, Haowei Zhang, Junxiao Song, Ruoyu Zhang, Runxin Xu, Qihao Zhu, Shirong Ma, Peiyi Wang, Xiao Bi, et~al.
\newblock Deepseek-r1: Incentivizing reasoning capability in llms via reinforcement learning.
\newblock {\em arXiv preprint arXiv:2501.12948}, 2025.

\bibitem{du2025virgo}
Yifan Du, Zikang Liu, Yifan Li, Wayne~Xin Zhao, Yuqi Huo, Bingning Wang, Weipeng Chen, Zheng Liu, Zhongyuan Wang, and Ji-Rong Wen.
\newblock Virgo: A preliminary exploration on reproducing o1-like mllm.
\newblock {\em arXiv preprint arXiv:2501.01904}, 2025.

\bibitem{zhou2025r1}
Hengguang Zhou, Xirui Li, Ruochen Wang, Minhao Cheng, Tianyi Zhou, and Cho-Jui Hsieh.
\newblock R1-zero's" aha moment" in visual reasoning on a 2b non-sft model.
\newblock {\em arXiv preprint arXiv:2503.05132}, 2025.

\bibitem{liu2025seg}
Yuqi Liu, Bohao Peng, Zhisheng Zhong, Zihao Yue, Fanbin Lu, Bei Yu, and Jiaya Jia.
\newblock Seg-zero: Reasoning-chain guided segmentation via cognitive reinforcement.
\newblock {\em arXiv preprint arXiv:2503.06520}, 2025.

\bibitem{zhan2025vision}
Yufei Zhan, Yousong Zhu, Shurong Zheng, Hongyin Zhao, Fan Yang, Ming Tang, and Jinqiao Wang.
\newblock Vision-r1: Evolving human-free alignment in large vision-language models via vision-guided reinforcement learning.
\newblock {\em arXiv preprint arXiv:2503.18013}, 2025.

\bibitem{deng2025boosting}
Huilin Deng, Ding Zou, Rui Ma, Hongchen Luo, Yang Cao, and Yu~Kang.
\newblock Boosting the generalization and reasoning of vision language models with curriculum reinforcement learning.
\newblock {\em arXiv preprint arXiv:2503.07065}, 2025.

\bibitem{peng2025lmm}
Yingzhe Peng, Gongrui Zhang, Miaosen Zhang, Zhiyuan You, Jie Liu, Qipeng Zhu, Kai Yang, Xingzhong Xu, Xin Geng, and Xu~Yang.
\newblock Lmm-r1: Empowering 3b lmms with strong reasoning abilities through two-stage rule-based rl.
\newblock {\em arXiv preprint arXiv:2503.07536}, 2025.

\bibitem{liu2025visual}
Ziyu Liu, Zeyi Sun, Yuhang Zang, Xiaoyi Dong, Yuhang Cao, Haodong Duan, Dahua Lin, and Jiaqi Wang.
\newblock Visual-rft: Visual reinforcement fine-tuning.
\newblock {\em arXiv preprint arXiv:2503.01785}, 2025.

\bibitem{yang2025r1}
Yi~Yang, Xiaoxuan He, Hongkun Pan, Xiyan Jiang, Yan Deng, Xingtao Yang, Haoyu Lu, Dacheng Yin, Fengyun Rao, Minfeng Zhu, et~al.
\newblock R1-onevision: Advancing generalized multimodal reasoning through cross-modal formalization.
\newblock {\em arXiv preprint arXiv:2503.10615}, 2025.

\bibitem{zhang2025r1}
Jingyi Zhang, Jiaxing Huang, Huanjin Yao, Shunyu Liu, Xikun Zhang, Shijian Lu, and Dacheng Tao.
\newblock R1-vl: Learning to reason with multimodal large language models via step-wise group relative policy optimization.
\newblock {\em arXiv preprint arXiv:2503.12937}, 2025.

\bibitem{deng2025openvlthinker}
Yihe Deng, Hritik Bansal, Fan Yin, Nanyun Peng, Wei Wang, and Kai-Wei Chang.
\newblock Openvlthinker: An early exploration to complex vision-language reasoning via iterative self-improvement.
\newblock {\em arXiv preprint arXiv:2503.17352}, 2025.

\bibitem{Alayrac2022}
Jean-Baptiste Alayrac et~al.
\newblock Flamingo: a visual language model for few-shot learning.
\newblock In {\em Advances in Neural Information Processing Systems}, volume~35, pages 23716--23736, 2022.

\bibitem{Liu2024b}
Haotian Liu et~al.
\newblock Visual instruction tuning.
\newblock In {\em Advances in Neural Information Processing Systems}, volume~36, 2024.

\bibitem{Achiam2023}
Josh Achiam et~al.
\newblock Gpt-4 technical report.
\newblock {\em arXiv preprint arXiv:2303.08774}, 2023.

\bibitem{Dai2023}
Wenliang Dai et~al.
\newblock Instructblip: Towards general-purpose vision-language models with instruction tuning.
\newblock In {\em Proceedings of the Conference on Neural Information Processing Systems}, 2023.

\bibitem{Chen2024a}
Kezhen Chen et~al.
\newblock Dragonfly: Multi-resolution zoom supercharges large visual-language model.
\newblock {\em arXiv preprint arXiv:2406.00977}, 2024.

\bibitem{Chen2024c}
Zhe Chen et~al.
\newblock How far are we to gpt-4v? closing the gap to commercial multimodal models with open-source suites.
\newblock {\em arXiv preprint arXiv:2404.16821}, 2024.

\bibitem{Li2024a}
Bo~Li et~al.
\newblock Llava-onevision: Easy visual task transfer.
\newblock {\em arXiv preprint arXiv:2408.03326}, 2024.

\bibitem{Bai2025a}
Shuai Bai et~al.
\newblock Qwen2.5-vl technical report.
\newblock {\em arXiv preprint arXiv:2502.13923}, 2025.

\bibitem{Wang2024a}
Peng Wang et~al.
\newblock Qwen2-vl: Enhancing vision-language model’s perception of the world at any resolution.
\newblock {\em arXiv preprint arXiv:2409.12191}, 2024.

\bibitem{Chen2024b}
Zhe Chen et~al.
\newblock Expanding performance boundaries of open-source multimodal models with model, data, and testtime scaling.
\newblock {\em arXiv preprint arXiv:2412.05271}, 2024.

\bibitem{Guo2025}
Daya Guo et~al.
\newblock Deepseek-r1: Incentivizing reasoning capability in llms via reinforcement learning.
\newblock {\em arXiv preprint arXiv:2501.12948}, 2025.

\bibitem{Yang2025}
Yi~Yang et~al.
\newblock R1-onevision: Advancing generalized multimodal reasoning through cross-modal formalization.
\newblock {\em arXiv preprint arXiv:2503.10615}, 2025.

\bibitem{Chen2025a}
Liang Chen et~al.
\newblock R1-v: Reinforcing super generalization ability in visionlanguage models with less than \$3.
\newblock \url{https://github.com/Deep-Agent/R1-V}, 2025.
\newblock Accessed: 2025-02-02.

\bibitem{Shao2024}
Zhihong Shao et~al.
\newblock Deepseekmath: Pushing the limits of mathematical reasoning in open language models.
\newblock {\em arXiv preprint arXiv:2402.03300}, 2024.

\bibitem{Zhou2025}
Hengguang Zhou et~al.
\newblock R1-zero’s” aha moment” in visual reasoning on a 2b non-sft model.
\newblock {\em arXiv preprint arXiv:2503.05132}, 2025.

\bibitem{fangtpami2021}
Deng-Ping Fan, Ge-Peng Ji, Ming-Ming Cheng, and Ling Shao.
\newblock Concealed object detection.
\newblock {\em IEEE TPAMI}, 2021.

\bibitem{hu2023high}
Xiaobin Hu, Shuo Wang, Xuebin Qin, Hang Dai, Wenqi Ren, Donghao Luo, Ying Tai, and Ling Shao.
\newblock High-resolution iterative feedback network for camouflaged object detection.
\newblock In {\em Proceedings of the AAAI Conference on Artificial Intelligence}, volume~37, pages 881--889, 2023.

\bibitem{tang2024chain}
Lv~Tang, Peng-Tao Jiang, Zhi-Hao Shen, Hao Zhang, Jin-Wei Chen, and Bo~Li.
\newblock Chain of visual perception: Harnessing multimodal large language models for zero-shot camouflaged object detection.
\newblock In {\em Proceedings of the 32nd ACM International Conference on Multimedia}, pages 8805--8814, 2024.

\bibitem{cuthill2019camouflage}
IC~Cuthill.
\newblock Camouflage.
\newblock {\em JOZ}, 308(2):75--92, 2019.

\bibitem{stevens2009animal}
Martin Stevens and Sami Merilaita.
\newblock Animal camouflage: current issues and new perspectives.
\newblock {\em PTR:BS}, 364(1516):423--427, 2009.

\bibitem{shao2024deepseekmath}
Zhihong Shao, Peiyi Wang, Qihao Zhu, Runxin Xu, Junxiao Song, Xiao Bi, Haowei Zhang, Mingchuan Zhang, YK~Li, Y~Wu, et~al.
\newblock Deepseekmath: Pushing the limits of mathematical reasoning in open language models.
\newblock {\em arXiv preprint arXiv:2402.03300}, 2024.

\bibitem{bai2025qwen2}
Shuai Bai, Keqin Chen, Xuejing Liu, Jialin Wang, Wenbin Ge, Sibo Song, Kai Dang, Peng Wang, Shijie Wang, Jun Tang, et~al.
\newblock Qwen2. 5-vl technical report.
\newblock {\em arXiv preprint arXiv:2502.13923}, 2025.

\bibitem{fan2020camouflaged}
Deng-Ping Fan, Ge-Peng Ji, Guolei Sun, Ming-Ming Cheng, Jianbing Shen, and Ling Shao.
\newblock Camouflaged object detection.
\newblock In {\em CVPR}, 2020.

\bibitem{lv2021simultaneously}
Yunqiu Lv, Jing Zhang, Yuchao Dai, Aixuan Li, Bowen Liu, Nick Barnes, and Deng-Ping Fan.
\newblock Simultaneously localize, segment and rank the camouflaged objects.
\newblock In {\em CVPR}, 2021.

\bibitem{le2019anabranch}
Trung-Nghia Le, Tam~V Nguyen, Zhongliang Nie, Minh-Triet Tran, and Akihiro Sugimoto.
\newblock Anabranch network for camouflaged object segmentation.
\newblock {\em CVIU}, 184:45--56, 2019.

\bibitem{2018Animal}
P~Skurowski, H~Abdulameer, J~Błaszczyk, T~Depta, A~Kornacki, and P~Kozieł.
\newblock Animal camouflage analysis: Chameleon database.
\newblock Unpublished Manuscript, 2018.

\end{thebibliography}
}

\end{document}